\documentclass[letterpaper, 10 pt, conference]{ieeeconf}  

\IEEEoverridecommandlockouts                              

\newtheorem{mydef}{Definition}
\usepackage{alltt}
\usepackage{subcaption}
\usepackage{mathtools}
\usepackage{multirow}
\usepackage{float}
\DeclareMathOperator*{\argmax}{arg\,max}
\DeclareMathOperator*{\argmin}{arg\,min}

\title{\LARGE \bf
Order Matters: Generating Progressive Explanations for \\Planning Tasks in Human-Robot Teaming
}

\author{Mehrdad Zakershahrak, Shashank Rao Marpally, Akshay Sharma, Ze Gong, and Yu Zhang
\thanks{Mehrdad Zakershahrak, Shashank Rao Marpally, Akshay Sharma, Ze Gong, and Yu Zhang are with the School of Computing, Informatics and Decision Systems Engineering, Arizona State University, Tempe, AZ.
        {\tt\small \{mzakersh, smarpall, ashar204, zgong11, yu.Zhang.442\}@asu.edu} 
         {}}%
}

\begin{document}

\maketitle
\thispagestyle{empty}
\pagestyle{empty}

\begin{abstract}
Prior work on generating explanations in a planning and decision-making context has focused on providing the rationale behind an AI agent's decision making. While these methods provide the right explanations from the explainer's perspective, they fail to heed the cognitive requirement of understanding an explanation from the explainee's (the human's) perspective.
In this work, we set out to address this issue by first considering the influence of information order in an explanation, or the {\it{\textbf{ progressiveness of explanations}}}. 
Intuitively, progression builds later concepts on previous ones and is known to contribute to better learning. 
In this work, we aim to investigate similar effects during explanation generation when an explanation is broken into multiple parts that are communicated sequentially.
The challenge here lies in modeling the humans' preferences for information order in receiving such explanations to assist understanding.
Given this sequential process, a  formulation based on goal-based MDP for generating progressive explanations is presented. 
The reward function of this MDP is learned via inverse reinforcement learning based on explanations that are retrieved via human subject studies.
We first evaluated our approach on a scavenger-hunt domain to demonstrate its effectively in capturing the humans' preferences. 
Upon analyzing the results, it revealed something more fundamental: 
the preferences arise strongly from both domain dependent and independence features. 
The correlation with domain independent features pushed us to verify this result further in an escape room domain.  
Results confirmed our hypothesis that the process of understanding an explanation was a dynamic process.  
The human preference that reflected this aspect 
corresponded exactly to the progression for knowledge assimilation hidden deeper in our cognitive process.  
We showed that progressive explanations achieved better task performance and reduced cognitive load.
These results sheds light on designing explainable robots across various domains. 

\end{abstract}

\section{INTRODUCTION}

As robots start to benefit a diverse set of domains, human-robot interaction has evolved to be an increasingly important subject.
In human-robot teaming, in particular, it is desired that the interaction occurs 
in a coherent manner that is observed in human-human teaming \cite{chakraborti2017ai,cooke2015team}. Similar to a human teammate, a robotic agent is required to not only understand its human partners, but also explain its own decisions or behaviors when necessary. 
Explanations in a teaming context provide the rationale behind an individual agent's decision making \cite{lombrozo2006structure}, and help with building a shared situation awareness and maintaining trust between teammates \cite{endsley1988design,cooke2015team}.
Although there exists prior work on generating explanations, the focus has been on generating the right explanations from the explainer's perspective rather than good explanations for the explainee \cite{gobelbecker2010coming,hanheide2017robot,sohrabi2011preferred}. 

Unsurprisingly, the right explanation may not necessarily be a good explanation--anyone with teaching or mentoring experience would share the sympathy. Such dissonance between the explainer and explainee may be a result of various inconsistencies, such as information asymmetry or different cognitive capabilities, just to name a few. These inconsistencies may be summarized as \textit{model differences}--the differences between the cognitive models that govern the generation and interpretation of an explanation, respectively, for the explainer and explainee~\cite{chakraborti2019explicability}. When these two models are the same, as is assumed in most prior works, an explanation from the perspective of the explainer would be not only correct but also perfectly understandable to the explainee, that is, as if the explanation were made to the explainer himself.
The more general case when the models differ has also been investigated \cite{chakraborti1802plan,zakershahrak2019online} under the model reconciliation setting, where the focus is on explaining domain model differences such that the two models become more compatible.
One remaining challenge in explanation generation, however, is to account for the differences in the cognitive capabilities to understand an explanation. 


In this work, we take a step further by generating explanations while considering the differences between the cognitive capabilities of the explainer and explainee. This is especially relevant to human-robot teaming since robots are frequently deployed to situations that require high cognitive and computational abilities that we do not have. 
To accommodate this, 
the motivation here is to generate explanations that reduce the cognitive effort required to understand them for the explainee.
In this work, we focus on the influence of the order of information.
In a moderately complex domain, making an explanation is not an instantaneous effort; instead, information must be conveyed in small parts sequentially.
Our proposal in studying the order of information is inspired by studies in psychology and education
on the limitations of human cognitive systems~\cite{ericsson1991toward,kahneman2011thinking} 
and progression in learning~\cite{schwarz2009developing}. 
Consequently, we term our approach \textit{progressive explanation generation}. Consider the following example of a conversation between two friends, which illustrates the importance of providing information in a proper order when making an explanation:

{\small{
\begin{quote}
\begin{verbatim}
Amy: Let's go to the outlet today.
Monica: My car is ready.
Amy: Great!
Monica: The rain will stop soon.
Amy: Wonderful!
Monica: By the way, today is a holiday
(shops closed).
Amy: You are telling me now!
Monica: Let us go to the central park!
Amy: ...
\end{verbatim}
\end{quote}
}}

Such cognitive dissonance illustrated above occurs frequently in our lives and it is our aim in this work to avoid similar situations when a robot is making an explanation to you.
The challenge lies in modeling the  humans'  preferences  for  information  order  in  receiving such explanations to assist understanding.
To this end, a general formulation based on goal-based Markov Decision Processes for generating progressive explanation is presented given the sequential information communication in an explanation. 
We propose to learn a quantification of the cognitive effort for each step as a reward function in an inverse reinforcement framework~\cite{ng2000algorithms,abbeel2004apprenticeship,ziebart2008maximum}. 
We set out to validate the following hypothesis:

\begin{itemize}
    \item H1. Our learning method can learn about the humans' preferences in receiving explanations.
\end{itemize}

Both domain-dependent and domain-independent features are used in learning based on explanations provided via human subject studies.
We evaluated first on a scavenger-hunt domain. 
Upon  analyzing  the  results,  however, it  revealed something  more  fundamental:  the  preferences  arise  strongly from  both  domain  dependent  and  independence  features.  The correlation  with  domain  independent  features  pushed  us  to verify  this  result  further  in  an  escape  room  domain.  
The strong weights on domain independent features, 
which capture plan changes during the explanation process,
implies that understanding an explanation is a dynamical process:

\begin{itemize}
    \item H2. Humans replan {\textit{\textbf{dynamically}}} to understand during an explanation instead of after an explanation for moderately and highly complex tasks. 
\end{itemize}

 Results confirmed our hypothesis that the process of understanding an explanation was a dynamic process. The human preference that reflected this aspect corresponded exactly to the progression for knowledge assimilation hidden deeper in our cognitive process.
Our results will benefit the design of robots that make explanations across various domains since such a preference is domain independent.
The last hypothesis is about the effectiveness of progressive explanations:
\begin{itemize}
    \item H3: Progressive explanations reduces cognitive load and improves task performance. 
\end{itemize}

Comparison with two baseline methods validated H3. 
We showed that the progressiveness in explanations corresponded well to the ``progressiveness'' of the curve on domain independent features.




\section{Related Work}

Explainable AI \cite{gunning2017explainable} is increasingly considered to be an important paradigm for designing future intelligent agents, especially as such systems begin to constitute an important part of our lives. The key requirement of explainable agency \cite{langley2017explainable} is to be ``{\it explainable}'' to the human partners. To be explainable, an agent must not only provide a solution to achieve a goal, but also make sure that the solution is perceived as such by its human peers. A determinant here is the human's interpretation of the agent's behavior. It is critical to take careful steps to avoid situations where the agent's assistance would be interpreted as no more than an interruption, which resulted in the pitfall of earlier effort in designing intelligent assistants, such as the loss of situation awareness and trust \cite{endsley2016designing,langfred2004too}.

The key challenge to explainable agency hence is the ability to model the human cognitive model that is responsible for interpreting the behaviors of other agents \cite{chakraborti2017ai}. With such a model, there are different ways to make the robot's behavior explainable. One way is to bias the robot's behavior towards the human's expectation of it based on the human's cognitive model. Under this framework, a robot can generate legible motions \cite{dragan2013generating} or explicable plans \cite{zhang2017plan,zakershahrak2018interactive}. Essentially, the robot sacrifices the plan quality to respect the human's expectation--the resulting plan is often a more costly plan. Another way is to provide a forewarning of the robot's intention before execution, such as for persuasion~\cite{petty1979effects}. In~\cite{gong2018robot}, the approach there is to provide additional context to help explain the robot's decision. The third way, which is the most relevant to ours, is for the robot to explain its decision via explanations \cite{gobelbecker2010coming,hanheide2017robot,sohrabi2011preferred}. The benefit of generating explanations, compared to generating explainable plans, is that the robot can keep its original (and optimal) plan. However, as mentioned earlier, the focus there is often on providing the rationale behind the explainer's decision making, while largely ignoring the explainee. In \cite{chakraborti1802plan}, this gap is addressed by considering  explanation generation as a model reconciliation problem, which takes into account the explainee's model. 
Although the cognitive requirement is implicitly considered, the  aim there is to reconcile (i.e., reduce) the differences in domain models, so that the robot's plan would be interpretable also in the model of the explainee.

The idea of progressive explanation generation
is connected to heuristics in the planning literature for reducing replanning effort~\cite{fox2006plan}. 
However, the focus there is on system effort~\cite{likhachev2005anytime}.
Our results, however, suggest that these heuristics may have a deeper connection to the human cognitive process in decision-making domains.

\section{Model Reconciliation}
\begin{figure} 
    \centering
    \includegraphics[width = 0.8\columnwidth]{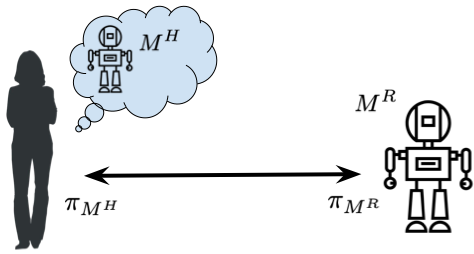}
    \caption{Explanation generation as model reconciliation \protect\cite{chakraborti1802plan}. $M^R$ denotes the robot model and $M^H$ denotes the human model that is used to generate her expectation of the robot's behavior ($\pi_{M^H}$). When the expectation does not match the robot's behavior, $\pi_{M^R}$, explanations must be generated.}
    \label{fig:setting}
    \vspace{-5mm}
\end{figure}

We base our work on a general model reconciliation setting for explanation generation that considers both the models of the explainer and explainee, which is introduced in~\cite{chakraborti1802plan}.
As shown in Fig. \ref{fig:setting}, the human uses $M^H$ to generate her expectation of the robot's behavior, while the robot's actual behavior is being created using the robot's model $M^R$, which is different from $M^H$. Therefore, $\pi_{M^R}$, which is the plan created from $M^R$, could also be different with $\pi_{M^H}$, which is the plan created from $M^H$. Whenever these two plans differ, the robot's plan must be explained. 

\begin{mydef}[Model Reconciliation~\cite{chakraborti1802plan}] \label{def:recon}
A model reconciliation setting is a tuple $(\pi^\ast_{I,G}, \langle M^R, M^H \rangle)$, where $cost(\pi^\ast_{I,G}, M^R) = cost^{\ast}_{M^R}(I,G)$.
\end{mydef}
\noindent where $\pi^\ast_{I,G}$ is the robot's optimal plan to be explained. $cost(\pi^\ast_{I,G}, M^R)$ is the cost of the robot's plan under the model $M^R$. $cost^{\ast}_{M^R}(I,G)$ returns the optimal plan given the initial state and goal state pair using $M^R$. Therefore, the constraint in the Definition \ref{def:recon} ensures that the robot's plan is optimal in its own model.


In this setting, the robot must generate an explanation to modify the human's model $M^H$ such that $\pi^\ast_{I,G}$ becomes explainable in the human's modified model (denoted as $\widehat{M^H}$) after the reconciliation. 
As a result, an explanation for a model reconciliation setting can be considered as requesting changes to the model of the human. Note that making an explanation may also lead to an error report if it is identified that the robot's model was incorrect.

To capture the model changes, a model function $\Gamma: \mathcal{M} \rightarrow 2^F$ is defined to convert a model to a set of model features \cite{chakraborti1802plan}, where $\mathcal{M}$ is the model space and $F$ the feature space. In this way, one model can be updated to another model with editing functions that change one feature at a time. The set of feature changes is denoted as $\Delta(M_1, M_2)$ and the distance between two models as the number of such feature changes is denoted as $\delta(M_1, M_2)$. In this work, we assume that the model is defined in PDDL \cite{fox2003pddl2}, an extension of STRIPS~\cite{fikes1971strips}, where a model is specified as a tuple $M = (D, I, G)$. The domain $D = (F,A)$ is comprised of a set of predicates, $F$, and a set of actions, $A$. $F$ is used to specify the state of the world. Each action $a \in A$ changes the state of the world by adding or deleting predicates. Therefore, an action can be represented as $a = (pre(a), eff^{+}(a), eff^{-}(a), c_a)$; where $pre(a)$ denotes the preconditions of the action $a$, and $eff^{+}(a), eff^{-}(a)$ indicate add and delete effects, respectively, and $c_a$ is the cost of the action. For example, a very simple model for Amy in our motivating example would be:

{\small{
\begin{alltt}
\textbf{Initial state:} not-holiday
\textbf{Goal state:} happy
\textbf{Actions:}
OUTLET-SHOPPING 
\hskip5pt pre: not-holiday (car-ready is-sunny)
\hskip5pt eff\textsuperscript{+}: happy
VISIT-PARK 
\hskip5pt pre: (car-ready is-sunny)
\hskip5pt eff\textsuperscript{+}: happy
\end{alltt}
}}

\normalsize
For simplicity, we use only boolean variables above.
The variables in parenthesis are optional predicates that are preferred but not required. 
The goal is to achieve the effect of \texttt{happy}. In this example, the model, denoted as $M_{\texttt{Amy}}$, will be converted by the model function $\Gamma$ to:

{\small{
\begin{quote}
\begin{alltt}
\(\Gamma(M_\texttt{Amy})\) = \{
    init-has-not-holiday,
    goal-has-happy,
    OS-has-precondition-not-holiday,
    OS-has-add-effect-happy, \dots\}
\end{alltt}
\end{quote}
}}
\normalsize
where $OS$ above is short for \texttt{OUTLET-SHOPPING}. The function essentially turns a model into a set of features that fully specifies the model. Hence, changing the set of features will also change the model.

\begin{mydef}[Explanation Generation] \label{def:gen}
The explanation generation \cite{chakraborti1802plan} problem is a tuple $(\pi^\ast_{I,G},\langle M^R$, $M^H \rangle)$ where an explanation is a subset of $\Delta(M_R, M_H)$ such that: \\ 1) $\Gamma(\widehat{M^H}) \setminus \Gamma(M^H) \subseteq \Gamma(M^R)$, and 
\\ 2) $cost(\pi^\ast_{I,G}, \widehat{M^H}) - cost^{\ast}_{\widehat{M^H}}(I,G) < cost(\pi^\ast_{I,G}, M^H) - cost^{\ast}_{M^H}(I,G)$.
\end{mydef}
\noindent where $\widehat{M^H}$ denotes the model after the changes. The first condition requires the changes to the human's model to be consistent with the robot's model. 
The second condition states that the robot's plan must be closer (in terms of cost) to the optimal plan after the model changes than the situation before--an explanation should have the effect of moving the expected plan closer to the robot's optimal plan. 

\begin{mydef}[Complete Explanation]
A complete explanation \cite{chakraborti1802plan} is an explanation that additionally satisfies $cost(\pi^\ast_{I,G}, \widehat{M^H}) = cost^{\ast}_{\widehat{M^H}}(I,G)$.
\end{mydef}

A complete explanation requires the model changes to make the robot's plan also optimal in the changed human model, so that the robot's plan becomes interpretable in the human's model as well.
A minimally complete explanation (MCE) is also defined in~\cite{chakraborti1802plan}, which is a complete explanation with the minimum number of unit feature changes. 
An example of $\widehat{M_{\texttt{Amy}}}$ (corresponds to $\widehat{M^H}$) after a minimally complete explanation is:

{\small{
\begin{alltt}
\textbf{Initial state:}
not-holiday car-ready (+) is-sunny (+)
\textbf{Goal state:} happy
\textbf{Actions:}
OUTLET-SHOPPING
pre: not-holiday (car-ready is-sunny)
eff+: happy
VISIT-PARK
pre: (car-ready is-sunny)
eff+: happy
\end{alltt}
}}
\normalsize
where the strikeout denotes the feature removed and $+$'s denote additions. These changes correspond to the explanation made in our motivating example. In this case, the robot model, $M^R$, corresponds to $M_{\texttt{Monica}}$, is the same as $\widehat{M_{\texttt{Amy}}}$ after the explanation (with the model changes incorporated).

\section{Progressive Explanation Generation}
In progressive explanation generation, our focus is on how the ordering of presenting information in an explanation may affect its understanding. 
An explanation in our setting is naturally specified as a sequence of feature changes. 
Since we process information as it is received, the cumulative cognitive effort can then be considered as the sum of effort associated with understanding each feature change in a sequential order. 
We couple the cognitive effort for each change with a {\it model distance metric}, denoted as $\rho(M_i, M_{i+1})$ for the $i$th feature change, where $M_{i}$ is the model before the $i$-th feature change and $M_{i+1}$ is the model after that change. Thereby, progressive explanation generation can be defined as the following optimization problem:

\begin{mydef} [Progressive Explanation Generation (PEG)] \label{def:PEG}
A progressive explanation is a complete explanation with an ordered sequence of unit feature changes that minimize the sum of the model distance metric: $\argmin_{\Delta(\widehat{M^H}, M^H)} \sum_{f_i \in \langle \Delta(\widehat{M^H}, M^H) \rangle} \rho_i$, where $\rho_i$ is short for $\rho(M_i, M_{i+1})$, $i$ is the index of unit feature changes, and $f_i$ denotes the $i$-th unit feature change. 
\end{mydef}

The angle brackets above convert a set to an ordered set and the summation is over every unit feature change in an explanation, which is computed for before and after each unit feature change is made in a progressive fashion. 
The goal of PEG is to minimize the cumulative model distance metric, and thereby minimize the cognitive effort required from the explainee to understand the explanation.

\subsection{Learning the Model Distance Metric}
In the previous section, we introduced different distant-metrics for the model-plan to provide the explanations. In this section, we want to extend our approach further, by learning a distance metric for PEG based on the human preferences in a human-robot teaming scheme.

To learn the model distance metric for PEG, we formulate the problem as an inverse reinforcement learning (IRL) \cite{ng2000algorithms,abbeel2004apprenticeship,ziebart2008maximum} framework, where we assume the task of generating explanations can be expressed as a goal-based Markov Decision Processes (MDP).
A goal-based MDP is defined by a 6-element tuple $(S, A, T, R, \gamma, G)$, where $S$ is the state space and $A$ is the action space. The domain dynamics is represented as the transition function $T$ that determines the probability of transitioning into state $s'$ when taking an action $a$ in state $s$ (i.e., $P(s'|s,a)$). $R$ is the reward function and the goal of the agent is to maximize the expected cumulative reward. $\gamma$ is the discount factor that encodes the agent's preference of current rewards over future rewards. $G$ is a set of goal states where 
for each $g \in G$, $T (g, a, g) = 1, \forall a \in A$. We chose goal-based MDP since in each scenario, although the start state could be the same, the goal could be different and therefore the policy would be different.

\begin{figure} 
    \centering
    \includegraphics[width=\columnwidth]{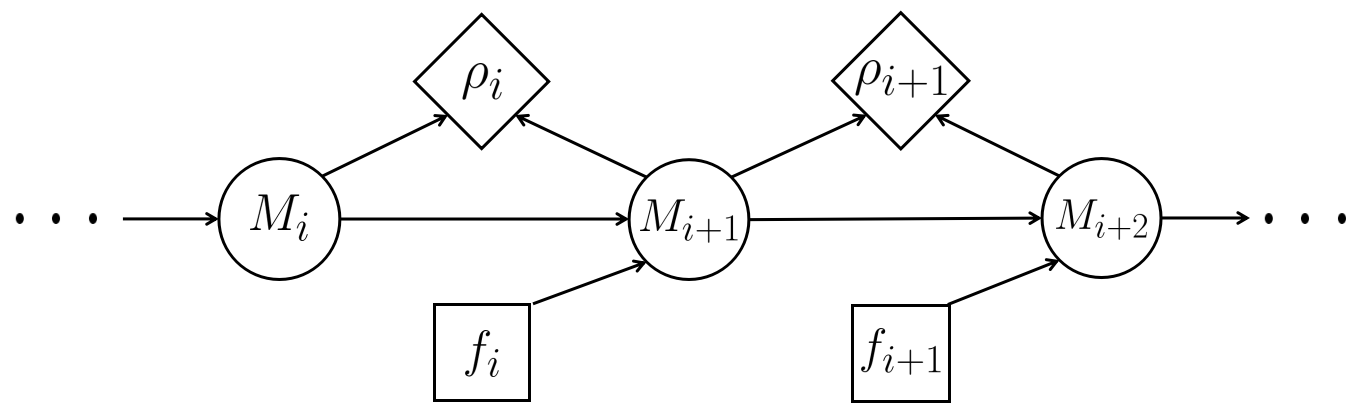}
    \caption{Illustration of the MDP that underlies PEG. At each time step, the human's model $M_i$ serves as the state. When the robot provides a unit feature change $f_i$ (as part of the explanation) to the human, the model changes according to $f_i$ to be the next state, $M_{i+1}$. The model distance metric $\rho_i$, which is short for $\rho(M_i,M_{i+1})$, captures the cognitive effort required to understand $f_i$.}
    \label{fig:fapx}
    \vspace{-5mm}
\end{figure}

Fig. \ref{fig:fapx} demonstrates the MDP that underlies PEG. In our work, the state space $S$ is the set of all possible human models and the action space $A$ is the set of all possible unit feature changes. The transition function $T$ captures the probability that the human model would be updated to $M'$ when the human model is $M$ and the robot explains $f$ to her (i.e., $P(M'|M,f)$). The model distance metric $\rho$ serves as the reward function, which depends on both the current and updated human models. 

\subsection{Applying IRL} \label{sec:irl}
Following prior work on IRL \cite{ng2000algorithms,abbeel2004apprenticeship,ziebart2008maximum}, we define the distance metric as a linear combination of a set of weighted features:
\begin{equation}
    \rho(M, M') = \sum_i \theta_i \cdot \psi_i(M, M') = \Theta^T \Psi(M, M') \nonumber
\end{equation}
where $\Psi=\{\psi_1, \psi_2, \dots, \psi_k\}$ is the set of features with respect to state pair $(M, M')$. $\Theta=\{\theta_1, \theta_2, \dots, \theta_k\}$ is the set of weights corresponding to the features.

Given a set of traces in a domain as a set of explanations (each is a sequence of unit features changes), which are obtained from human subjects, our goal is to learn the model distance metric $\rho$, which in turn requires us to learn the weights $\Theta$ given a set of features. 
Since noise is expected in the traces, we learn the weights by maximizing the likelihood of the traces using MaxEnt-IRL \cite{ziebart2008maximum} as follows:

{\small{
\begin{equation}
\begin{split}
    \Theta^* = \argmax_{\Theta} \mathcal{L}(D) &= \argmax_{\Theta}\frac{1}{|D|}\log P(D|\Theta) \\
    &= \argmax_{\Theta}\frac{1}{|D|}\sum_{G \in \mathcal{G}} \sum_{\widehat{\zeta_G} \in D_G} \log P(\widehat{\zeta_G}|\Theta) \label{equ:optimize}
\end{split}
\end{equation}
}}

\normalsize
where $D$ is the training data set, $\mathcal{G}$ the collection of goal sets $G$ for different scenarios. $\widehat{\zeta_G}=(M_0, f_1, M_1, 
$ $\dots$, $f_{n}, M_n)$ is an explanation for achieving $G$ with ordered feature changes provided by human subjects in a subset $D_G$. 
It consists of the initial human model (i.e., $M_0=M^H$), unit feature change and the updated model at each time step. To mitigate the ambiguity that the distribution of the traces may introduce preference for some traces over others, the principle of maximum entropy \cite{ziebart2008maximum} is employed to define the distribution over all the possible traces for a specific goal (i.e., $G$):
\begin{equation}
    P(\zeta_G|\Theta) = \frac{e^{\rho(\zeta_G)}}{\sum_{\zeta_G} e^{\rho(\zeta_G)}} \label{equ:maxent} \\
\end{equation}
where 
\begin{equation}
    \rho(\zeta_G) = \Theta^T \Psi(\zeta_G) = \sum_{\mathclap{(M, M') \in \zeta_G}} \Theta^T \Psi(M,M') \nonumber 
\end{equation}
Take Equation \ref{equ:maxent} into Equation \ref{equ:optimize}, the optimization becomes:
\begin{equation}
\Theta^* = \argmax_{\Theta}\frac{1}{|D|}\sum_{G \in \mathcal{G}}\sum_{\widehat{\zeta_G} \in D_G} \Big(\Theta^T \Psi(\widehat{\zeta_G}) - \log \sum_{\zeta_G} e^{\Theta^T \Psi(\zeta_G)}\Big) \label{equ:convex}
\end{equation}
Note that $\widehat{\zeta_G} \in D_G$ in the first term above represents a trace in the training data set while $\zeta_G$ in the second term above refers to {\it any} possible trace of the domain.
Since Equation \ref{equ:convex} is convex,
we use a gradient-based method to learn $\Theta$ and divide the traces into pairs of human models as in \cite{ziebart2008maximum}:
\small
\begin{equation}
\begin{split}
    \nabla_{\Theta}\mathcal{L} = \frac{1}{|D|} \sum_{G \in \mathcal{G}}\Big(
    \sum_{\mathclap{\substack{ \\ \\ (M,M') \in D_{G}}}} \Psi(M,M')  - 
     \sum_{\mathclap{(M,M') \in D_{G}}}P(M,M'|\boldsymbol{\theta}) \Psi(M,M')\Big) \nonumber
\end{split}
\end{equation}
\normalsize
Different from traditional applications of MaxEnt-IRL~\cite{ziebart2008maximum}, the model distance metric in our work depends on both the current and next human model. As a result, $P(M,M'|\Theta)$ above represents the model pair occurrence frequency (MPOF) for a pair $(M,M')$, which can be computed using dynamic programming. If we denote the probability of occurrence of $(M,M')$ at time $t$ as $\mu_t(M,M')$, we then have $P(M,M'|\Theta) = \sum_t \mu_t(M,M')$. The updating rules for $\mu_t$ is as follows:
\begin{equation}
\begin{split}
    &\mu_1(M,M') = P \Big( (M_1, M_2) = (M, M') \Big) \\
    &\mu_{t+1}(M,M') = \sum_f \sum_{M''} \mu_t (M'', M) P(f|M) P(M'|M,f) \nonumber
\end{split}
\end{equation}
The values for $\mu_1$ are initialized to the probability of the state pair $(M,M')$ being the first pair of a trace. The probability of the occurrence of $(M,M')$ at a certain time step then is calculated based on the occurrence frequency of the previous state pair, which has $M$ as the second entry in the pair, any unit feature change $f$ that the robot would explain to the human while in state $M$ (i.e., according to a stochastic policy), and the probability that the human model would end up in $M'$ when explaining $f$ in state $M$ (i.e., the transition function). 

The stochastic policy $P(f|M)$ specifies the probability of explaining $f$ when the human model is $M$,
which is computed as 
$P(f|M)=\frac{P(M,f)}{P(M)}$. 
Similarly, they can be calculated using dynamic programming as in \cite{ziebart2008maximum}. $\mu_1 (M, M')$ can then be approximated using sampled traces generated by the stochastic policy and transition function in each iteration. After learning the parameters for the model distance metric, we utilize uniform cost search for a specific goal to retrieve the best sequence of $f_i$ from a common initial state by maximizing the reward of each state:
\begin{equation}
    \zeta_G^* = \argmax_{\zeta_G} \sum_{(M, M') \in \zeta_G} \Theta^T\Psi(M, M')
\end{equation}

\subsection{Features Selection}\label{sec:feature}
The features used in our learning algorithm for the model distance metric belong in general to two categories: domain dependent and domain independent features. 


The domain dependent features in our study are chosen to be the ones that we consider to have an impact on the cognitive load.
These features should be fully specifiable by $M$ and $M'$ only given our IRL formulation. 
Although this imposes a restriction on the set features we can select, 
it still allows for a rich set of possibilities for any given domain. 
In our future work, we will further investigate the impact of this restriction on the learned model distance metric.


Domain independent features are chosen to reflect replanning cost. 
We consider two types of domain independent features: $(1)$ action distance \cite{fox2006plan}, 
and $(2)$ cost distance. Each of them represents a type of plan distances. 
The motivation to use plan distances is that, as the information is communicated progressively for an explanation (as unit model changes), humans process it as it is received (i.e., replan based on the current information). Intuitively, the effort involved in the replanning process is correlated to how many changes must be made to a plan, which is often captured by a distance metric.
For any model $M_i$, we denote the plan as $\pi_i$.
The following distance metrics are considered:

\textbf{Action Distance:} The action distance feature represents the distance between two plans $\pi_i$ and $\pi_j$ obtained from states $M_i$ and $M_j$ respectively, as distance($\pi_i, \pi_j) = \dfrac{\sum_{k=1}^{n}|C(a_{ki}) - C(a_{kj})|}{max(cost(\pi_i),cost( \pi_j))}$. Where $n = |\pi_i \cup \pi_j|$ and $C(a_{ki})$ is the number of occurrences of action $a_k$ in plan $\pi_i$, and $cost(\pi_i)$ is the cost of plan $\pi_i$.

\textbf{Cost Distance:} Similarly, the cost distance is the difference between the cost of plans $\pi_i$ and $\pi_j$ obtained from $M_i$ and $M_j$ respectively: $C(\pi_i, \pi_j) = |cost^*_{M_i}(I,G) - cost^*_{M_j}(I,G)|$. 


\textbf{Levenshtein Distance:} The Levenshtein distance \cite{levenshtein1966binary} is the minimum editing distance between plans $\pi_i$ and $\pi_j$ obtained from $M_i$ and $M_j$. The larger the minimum editing distance, the more different the plans are. The equation below provides the mathematical definition used to calculate the Levenshtein distance:

\begin{small}
\[
  lav_{\pi_i,\pi_j}(m,n) = 
   \begin{cases} 
   \max(m,n) \qquad\qquad\quad \text{if } \min(m,n)=0, \\
   \min 
   \begin{cases}
   lav_{\pi_i,\pi_j}(m-1,n)+1 \\
   lav_{\pi_i,\pi_j}(m,n-1)+1 \enspace \text{otherwise.} \\
   lav_{\pi_i,\pi_j}(m-1,n-1)+\mathbf{1}_{\pi_i(m)\neq \pi_j(n)}\\
   \end{cases}
   \end{cases}
\]
 \end{small}
where $lav_{\pi_i,\pi_j}(m,n)$ is the distance between the first $m$ actions in $\pi_i$ and first $n$ actions in $\pi_j$. $\mathbf{1}_{\pi_i(m)\neq \pi_j(n)}$ returns $0$ when $\pi_i(m) = \pi_j(n)$ and returns $1$ otherwise. 

\section{Evaluation}
We evaluated our approach by conducting human-subject studies using Amazon Mechanical Turk (MTurk) in two different domains: scavenger-hunt and escape-room. These domains were designed to expose the subjects to moderately complex situations that required a non-trivial amount of cognitive effort in a short amount of time. 

\subsection{Scavenger-Hunt}
\begin{figure} [hbtp]
\centering
\includegraphics[scale=0.30]{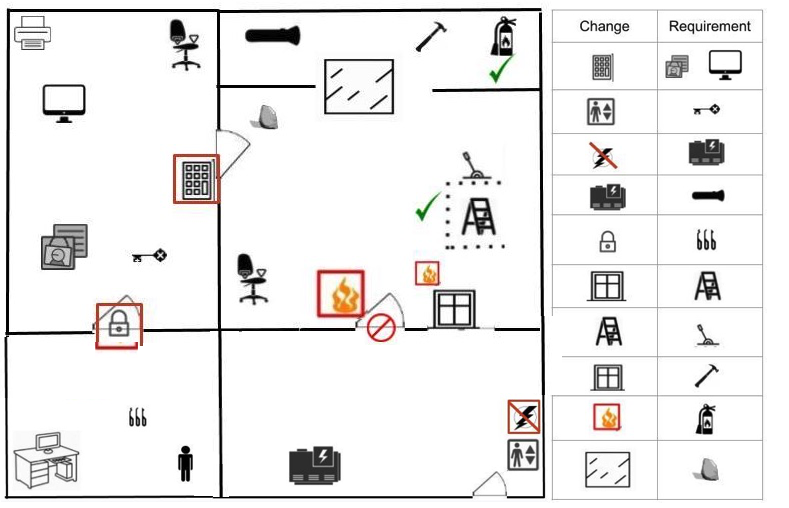}
\caption{Illustration of the scavenger-hunt domain.
}
\vspace{-4mm}
    \label{fig:scavenge}
\end{figure}

The task is situated in a damaged office building after an earthquake, portrayed by a floor-plan in Fig. \ref{fig:scavenge}. Normally, the human uses the doors connecting the rooms to exit the building via the elevator (bottom right corner) from his office (bottom left corner). However, an earthquake may interrupt the human's original path in different ways. 
The goal of the participant, as a member of the first response team, is to help the trapped human navigate through the building. 
At any step, the participant can explain one of the changes illustrated in a red box, such as a fire blocking the door or a power outage. 
The participants are given a map that contains all the possible changes. 
These are also shown on the right side of Fig. \ref{fig:scavenge}.
There are a total of $10$ possible changes that may have changed the plan of the trapped human. 
However, for any given scenario, 
only a few selected changes will be present. 


\subsubsection{Experiment Design}
We first explained domain to the participants and emphasized that whether the trapped person understood their explanations determined the life or death of that person. 
This was meant to encourage the participants to clearly explain the situation in a way to be understood. 
We then asked the participants to explain the situation to help the trapped human escape the building while playing the role of a member on the first response team. 
To ensure the quality of the data, We implemented a sanity check question to make sure the participants understood the task. We removed the responses with wrong answers to the  sanity questions or if it took them over 3 minutes to finish the task.

 

\begin{figure}[H]
    \centering
    \includegraphics[width=0.9\columnwidth]{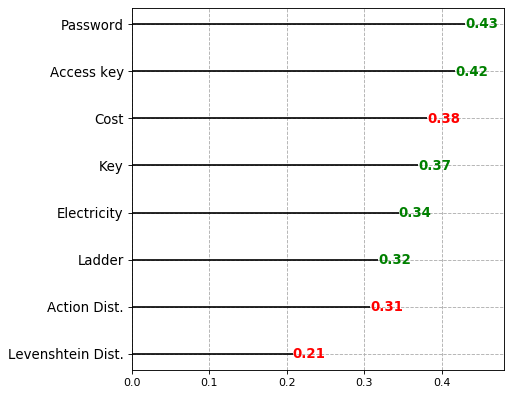}
    \caption{Normalized feature weights for the scavenger-hunt domain. The domain dependent features are one hot vector encoding for the state-pairs. 
    }
    \label{fig:ohv}
    \vspace{-4mm}
\end{figure} 

\subsubsection{Results \& Analysis}
We conducted a survey using Qualtrics and recruited $122$ human subjects using Mturk, with HIT acceptance rate of $99\%$. After sifting through the responses as described in the previous section, we used $93$ responses, out of which $66$ responses were over 5 training scenarios and $27$ responses were over 3 testing scenarios. 
The aim of this evaluation was to analyze if we could learn the human preferences from training scenarios and apply them to testing scenarios (H1).
We compared the outputs of our explanation generation algorithm based on the reward function learned by IRL with the subjects' responses for the testing scenarios. The accuracy of our method was $85.2\%$: our approach successfully matched $23$ out of the $27$ human responses across $3$ testing scenarios. 
This result showed that humans indeed had certain preferences for information order in such situations and that our method could capture these preferences.

Fig. \ref{fig:ohv} shows the weights learned by IRL for both domain dependent and domain independent features.
As Fig. \ref{fig:ohv} suggests, the domain independent features seem to have played an important role. 
This result inspired us further to verify since the significant weights on domain independent features, which captured plan changes during the explanation process, suggested that the understanding an explanation was a dynamics process. 
Consequently, we created another domain to verify this result and investigate further. 
This domain was introduced to impose similar preferences on different domain changes 
to minimize the influence of semantics (e.g., explaining a fire event could be naturally associated with a higher preference than obtaining a computer passcode).


\begin{figure} [hbtp]
\centering
\includegraphics[scale=0.4]{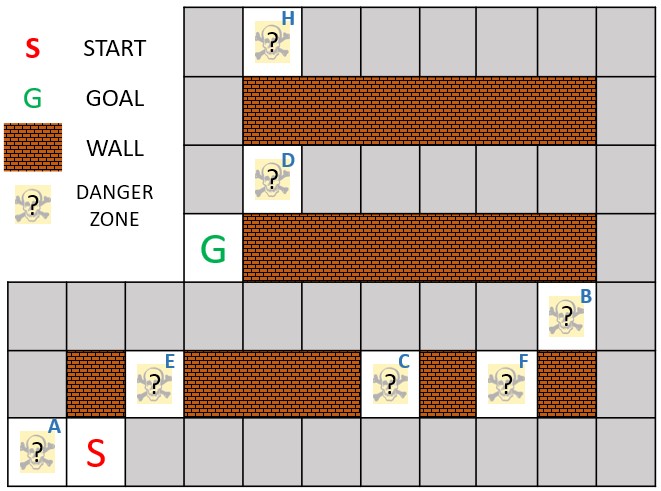}
\caption{Illustration of the escape-room domain.
}
    \label{fig:scenario}
    \vspace{-4mm}
\end{figure}

\subsection{Escape Room}
The task is situated in a damaged nuclear plant represented as a maze-like environment in Fig. \ref{fig:scenario}.
The goal of the human inside is to navigate from the starting location $S$ to the goal location $G$ as fast as as possible without going through dangerous locations. 
The set of actions in this domain are going to each of gateway cells from $S$, and then to $G$ from. For instance, \textit{go to cell $E$ from $S$, go to $G$ from $E$}, assuming cell $E$ is not a dangerous passage.
Some of the marked locations (see Fig. \ref{fig:scenario}) may be affected by the disaster, 
and become dangerous. Similarly, the participant played an external agent here to inform  the internal person about which locations were dangerous.  
The external agent could only convey one piece of information at a time (e.g., $D$ is a danger zone). 
The states of the $7$ marked locations correspond to $7$ contingencies (modeled as unit feature changes in the domain) that may have affected the human's plan. 

\subsubsection{Experimental Design} 
\label{sec:design}
We designed $8$ different scenarios for the escape-room domain.
We used $5$ scenarios for training and $3$ for testing.
Each scenario involves a different set of contingencies and we ensure that there are contingencies in the testing scenarios that did not appear in the training scenarios.
During training, 
the participants are at first introduced to the domain and informed that they are supposed to act as the external agent to communicate the contingencies to the internal person in the scenario. 
They are explained that the internal person is desperate to escape soon to give them a sense of urgency as well as an incentive to elucidate the situation. 
We also asked the participants at the beginning about what path the internal person would take assuming no marked locations are dangerous.
We use the answer to this question later to sift the data.


In this domain, we further introduced a testing phase for evaluating the effectiveness of PEG. 
In this phase, new participants played the role of the internal agent.
We tested the subject performance with our progressive explanation generation method and two baselines. In particular, we provided the subjects the contingencies that were ordered by 1) our method, 2) a random order, and 3) the Manhattan distance (of the contingency) relative to the starting location $S$.
To create a highly cognitive demanding situation, the subjects were pushed to  complete the task within 4 minutes.
Responses that failed the sanity check question or ran over 4 minutes were not used. 
After the task, the subjects were provided the NASA Task Load standard questionnaire (TLX) \cite{nasa} to evaluate the efficiency of the different methods.


\begin{figure}
    \centering
    \includegraphics[width=1.0\columnwidth]{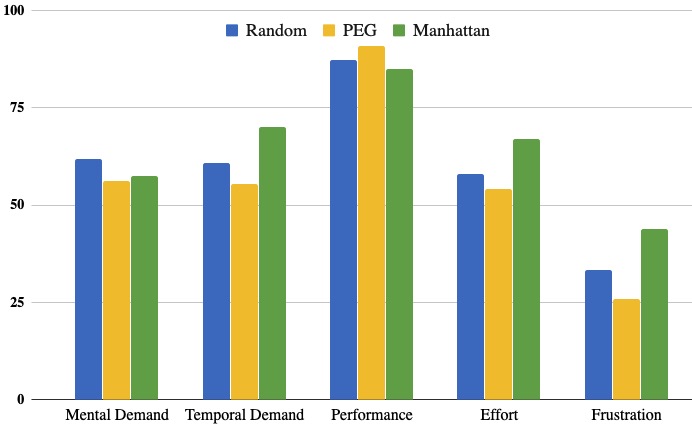}
    \caption{NASA TLX results for testing.}
    \label{fig:tlx}
    \vskip-10pt
\end{figure}

\subsubsection{Results \& Analysis} \label{sec:result}
To improve the quality of the responses, we set the criteria that the worker's HIT acceptance rate must be greater than $99\%$ and has been granted MTurk Masters. 
In the training phase, we created the surveys using Qualtrics and recruited $35$ human subjects on MTurk, out of which $21$ responses were used. 
For testing, we have recruited $163$ human subjects out of which $87$ responses were used. $58$ of our subjects were male and $29$ were female. The average of age of our subjects was $38.17$ with a standard deviation of $11.13$.
For domain dependent features, we chose $4$ features related to relative position of the contingency being explained with respect to the contingencies that have already been explained.
We refer to these features as $x_{min}$, $x_{max}$, $y_{max}$, and $y_{min}$.
Table \ref{tab:weight} shows the normalized weights $\Theta$ for each feature after learning via IRL as explained in Sec. \ref{sec:irl}.
Interestingly, the action distance and Levenshtein distance maintained high weights,
which was aligned with our prior results and further validated H2.
Simultaneously, however, the weight for the plan cost distance dropped significantly. 
We attributed this ``anomaly'' to the simple cost structure in the escape-room domain since
the increasing of plan cost does not necessarily increase the cognitive effort there. 


\renewcommand{\arraystretch}{1.2}
\begin{table}[htbp]
\center
\small{
\begin{tabular}{l|l|r}
\hline
    Feature Category & \multicolumn{1}{c|}{Feature Name} & Weights \\
    \hline
    \multirow{3}{*}{Domain independent}& Action Distance & 0.44\\
    & Cost Distance & 0.04\\
    & Levenshtein Distance & 0.46 \\
    \hline
    \multirow{2}{*}{Domain dependent}& 
     $x_{min}$, $y_{min}$ & 0.38, 0.41\\
    &$x_{max}$, $y_{max}$ & 0.35, 0.39\\
    \hline
\end{tabular}
\caption{Normalized feature weights for escape-room domain}
\label{tab:weight}
}
\end{table}

The subjective results for testing are presented in Fig. \ref{fig:tlx}.
We can see that our method (PEG) performs better than the baselines for all NASA TLX metrics, a statistically significant difference was observed between PEG and other methods for a weighted sum of TLX metrics, as shown in Table \ref{tab:subjective}. Objective metrics further confirmed that our method improved task performance as presented in Table \ref{tab:accuracy} which represents the percentage in which the participants came up with the correct plan after the respective explanations. 
This result verifies H3.


\begin{figure}
\centering
    \includegraphics[scale=0.30]{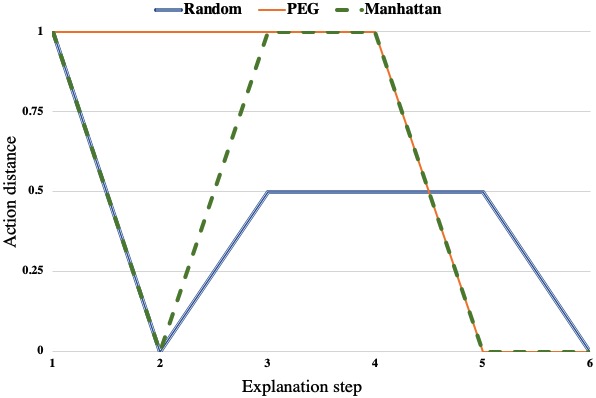}
    \caption{Changes of action distance per explanation step for escape-room domain}
    \label{fig:pcost}
\end{figure}

Fig. \ref{fig:pcost} shows the action distance per explanation step for one of the testing scenarios, which is very similar to the Levenshtein distance in this domain. An interesting observation is that the curve of PEG is smoother, i.e., it is missing the oscillation seen in other methods. 
This intuitively illustrates the progressiveness of explanation enabled by our method,
which suggests that the progressiveness of explanations is correlated with the progressiveness of these features.

\renewcommand{\arraystretch}{1.2}
\begin{table}
\setlength{\tabcolsep}{4pt}
 \scriptsize{
\centering
    \begin{tabular}{c|c|c|c|c|c|c}
    \hline
    ~         & ~ &  & \multirow{3}{*}{Performance} & \multirow{3}{*}{Effort} & \multirow{3}{*}{Frustration} & WT TLX \\
    ~ & Mental & Temporal & ~ & ~ & ~ & (exc \\
    ~ & Demand & Demand & ~ & ~ & ~ & Perf) \\ \hline
    Random    & 63.10         & 61.96           & 89.06       & 59.04  & 33.96       & 52.47                                \\ \hline
    \textbf{PEG}       & \textbf{56.19}     & \textbf{55.37}   & \textbf{90.74} & \textbf{54.11}  & \textbf{25.89}       & \textbf{41.93}                                \\ \hline
    Manhattan & 57.43         & 69.93           & 85.00       & 66.86  & 43.93       & 58.35                                \\ \hline
    \end{tabular}
\caption{Subjective results for each NASA TLX category}
\label{tab:subjective}}
\vskip-10pt
\end{table}

\begin{table} [htbp]
\footnotesize
\centering
    \begin{tabular}{l|c|c|c}
    \hline
    ~               & Random & PEG   & Manhattan \\ \hline
    Accuracy        & 85.4  (41/48) & \textbf{96.3  (26/27)} & 66.7  (8/12)  \\ \hline
    \end{tabular}
    \caption{Objective performance in terms of task accuracy}
    \label{tab:accuracy}
\end{table}

\section{Conclusions}
In this paper, we studied the problem of PEG.
We took a step further from the prior work 
by considering not only the right explanation for the explainee,
but also the underlying cognitive effort required to understand the explanation from the explainee's perspective,
resulting in a general framework for PEG. 
To address the challenge with modeling human preferences of the information order, 
we adopted the formulation of a goal-based MDP and applied IRL to learn the reward function based on traces. 
Our first contribution is that we show that humans indeed demonstrate preferences for the information order in explanations and we can indeed learn about such preferences using our framework. 
This verified H1. 
Upon analyzing the data, we noted strong weights for domain independent features, which suggested that the cognitive process for understanding an explanation is dynamic. 
This was validated in another domain. 
Together, results from these two domains validated H2. 
Finally, we showed that PEG did improve task performance and reduce cognitive load. 

Many future direction are possible. 
One interesting direction is to investigate sub-explanations as more than one unit feature change. 
This means that the robot may be allowed to explain multiple aspects at the same time.
One may anticipate that this would be useful  for aspects that are highly correlated. 
Another possible direction is to generalize the MDP model to remove the Markov assumption, 
which is  quite restrictive for modeling
human cognition.




\bibliographystyle{IEEEtran}
\bibliography{references}

\end{document}